\documentclass[preprint,12pt]{elsarticle}

\usepackage{amssymb}
\usepackage{amsmath}

\usepackage{booktabs}
\usepackage{hyperref}
\biboptions{sort&compress}
\usepackage{graphicx}
\usepackage[table,xcdraw]{xcolor}
\usepackage{multirow}

\DeclareMathOperator*{\argmax}{arg\,max}

\let\today\relax
\makeatletter
\def\ps@pprintTitle{%
    \let\@oddhead\@empty
    \let\@evenhead\@empty
    \def\@oddfoot{\footnotesize\itshape
         {Submitted preprint} \hfill\today}%
    \let\@evenfoot\@oddfoot
    }
\makeatother

\begin{document}

\begin{frontmatter}



\title{VisTA: Vision-Text Alignment Model with Contrastive Learning using Multimodal Data for Evidence-Driven, Reliable, and Explainable Alzheimer's Disease Diagnosis}


\author[chuv,unil,uet]{Duy-Cat Can} 
\ead{duy-cat.can@chuv.ch}
\author[hcmus]{Linh D. Dang} 
\ead{dang.diemlinh0212@gmail.com}
\author[hcmus]{Quang-Huy Tang} 
\ead{tqhuy23@apcs.fitus.edu.vn}
\author[mh175]{Dang Minh Ly} 
\ead{drminhdang@outlook.com}
\author[hcmiu]{Huong Ha} 
\ead{htthuong@hcmiu.edu.vn}
\author[chuv]{Guillaume Blanc} 
\ead{guillaume.e.blanc@gmail.com}
\author[chuv,unil]{Oliver Y. Ch\'en\corref{cor1}} 
\ead{olivery.chen@chuv.ch}
\author[hcmus]{Binh T. Nguyen\corref{cor1}} 
\ead{ngtbinh@hcmus.edu.vn}
\cortext[cor1]{Corresponding authors.}

\affiliation[chuv]{
    organization={Plateforme de bio-informatique, Centre hospitalier universitaire vaudois (CHUV)},
    addressline={Rue du Bugnon 46}, 
    city={Lausanne},
    postcode={1005}, 
    state={Vaud},
    country={Switzerland}
}
\affiliation[unil]{
    organization={Faculté de biologie et de médecine, Université de Lausanne (UNIL)},
    addressline={Quartier Centre}, 
    city={Lausanne},
    postcode={1015}, 
    state={Vaud},
    country={Switzerland}
}
\affiliation[uet]{
    organization={VNU University of Engineering and Technology, Vietnam National University},
    addressline={144 Xuan Thuy, Cau Giay}, 
    city={Hanoi},
    country={Vietnam}
}
\affiliation[hcmus]{
    organization={Department of Computer Science, University of Science, Vietnam National University Ho Chi Minh City},
    addressline={227 Nguyen Van Cu St., Ward 4, District 5}, 
    city={Ho Chi Minh City},
    country={Vietnam}
}
\affiliation[mh175]{
    organization={Department of Neurology, Department of Clinical Neurophysiology, Military Hospital 175},
    addressline={786 Nguyen Kiem Str, Ward 3, Go Vap District}, 
    city={Ho Chi Minh City},
    country={Vietnam}
}
\affiliation[hcmiu]{
    organization={School of Biomedical Engineering, International University, Vietnam National University},
    addressline={Quarter 6, Linh Trung Ward, Thu Duc City}, 
    city={Ho Chi Minh City},
    country={Vietnam}
}

\begin{abstract}
\textbf{Objective}:
Assessing Alzheimer's disease (AD) using high-dimensional radiology images is clinically important but challenging. Although Artificial Intelligence (AI) has advanced AD diagnosis, it remains unclear how to design AI models embracing predictability and explainability. Here, we propose VisTA, a multimodal language-vision model assisted by contrastive learning, to optimize disease prediction and evidence-based, interpretable explanations for clinical decision-making.

\textbf{Methods}:
We developed \textbf{VisTA} (Vision-Text Alignment Model) for AD diagnosis.
Architecturally, we built VisTA from BiomedCLIP and fine-tuned it using contrastive learning to align images with verified abnormalities and their descriptions.
To train VisTA, we used a constructed reference dataset containing images, abnormality types, and descriptions verified by medical experts.
VisTA produces four outputs: predicted abnormality type, similarity to reference cases, evidence-driven explanation, and final AD diagnoses.
To illustrate VisTA's efficacy, we reported accuracy metrics for abnormality retrieval and dementia prediction.
To demonstrate VisTA's explainability, we compared its explanations with human experts' explanations.

\textbf{Results}:
Compared to 15 million images used for baseline pretraining, VisTA only used 170 samples for fine-tuning and obtained significant improvement in abnormality retrieval and dementia prediction.
For abnormality retrieval, VisTA reached 74\% accuracy and an AUC of 0.87 (26\% and 0.74, respectively, from baseline models).  
For dementia prediction, VisTA achieved 88\% accuracy and an AUC of 0.82 (30\% and 0.57, respectively, from baseline models).  
The generated explanations agreed strongly with human experts' and provided insights into the diagnostic process.
Taken together, VisTA optimize prediction, clinical reasoning, and explanation.

\textbf{Conclusion}:
VisTA bridges the gap between black-box results and clinical demands by integrating evidence-driven reasoning and explainability.
VisTA enhances diagnostic reliability and aligns with real-world clinical workflows, offering a promising tool for automating AD diagnosis.
Future extensions of VisTA can include larger datasets and additional modalities beyond neuroimaging data to further improve generalizability.
\end{abstract}

\begin{graphicalabstract}
\textit{Corresponding authors:}\\
- Oliver Y. Ch\'en (olivery.chen@chuv.ch)\\
- Binh T. Nguyen (ngtbinh@hcmus.edu.vn)

\vspace{.5 mm}
\begin{center}
\includegraphics[width=\linewidth]{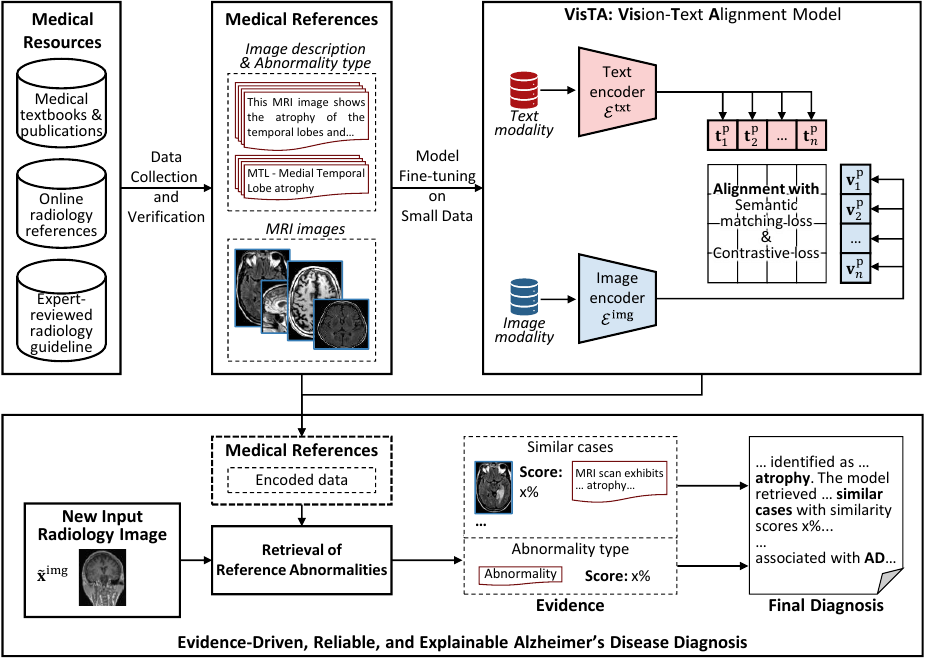}
\end{center}
\end{graphicalabstract}

\begin{highlights}
    \item We introduce \textbf{VisTA} (Vision-Text Alignment model), a multimodal machine learning framework for assessing Alzheimer's disease (AD) using radiology images with evidence-driven reasoning.
    
    \item We create the \textbf{MINDset} (Multimodal Imaging and NeuroDiagnostic dataset), a dataset consisting of radiology images, expert-verified abnormality types, and descriptive explanations, for developing and testing evidence-driven AI diagnostic.

    \item Using \textbf{contrastive learning}, VisTA only needs \textbf{170 samples} for fine-tuning to achieve significant improvements over pre-trained models using millions of images.
    
    \item VisTA includes a \textbf{modular diagnostic pipeline} that retrieves reference cases, aligns radiology images with verified abnormality descriptions, and generates interpretable predictions.
    
    \item VisTA shows outstanding \textbf{prediction performance} in abnormality retrieval (accuracy: 74\%; AUC: 0.87) and dementia prediction (accuracy: 88\%; AUC: 0.82) on MINDset, compared to baseline models.
    
    \item VisTA optimizes prediction with \textbf{explainability} and \textbf{reflects real-world clinical workflows}. It improves embedding space visualization, provides evidence-driven explanations, and enhances reliability during diagnostic and trust in AI-assisted decision-making.

    \item \textbf{Future directions} of VisTA include expanding the size of the data, incorporating additional data modalities, and improving generalizability for broader adoptation in healthcare.
\end{highlights}


\end{frontmatter}



\section{Introduction}
\subsection{Motivation}
Alzheimer's disease (AD) is a neurodegenerative disease that affects millions worldwide, posing significant health and economic challenges~\cite{who2023dementia, scheltens2021alzheimer, masters2015alzheimer}. An important way to deliver suitable AD management is to detect the disease early, timely, and accurately.
Radiology imaging, particularly magnetic resonance imaging (MRI), is a critical tool for detecting structural brain abnormalities associated with AD~\cite{jack2018nia, weiner2013alzheimer}.
While machine learning has advanced the prediction accuracy of AD using MRI scans~\cite{menagadevi2024machine, illakiya2023automatic, yao2023artificial}, interpreting these images, at present, requires extensive expert knowledge, may be biased due to disease variability, and prone to subjectivity~\cite{chow2016correlation}.
As clinical Artificial Intelligence (AI) advances, there is a pressing demand for building AI models that not only achieve accurate predictions but also provide transparent, explainable, and reliable results~\cite{holzinger2018machine, amann2020explainability, sadeghi2024review}.

Embracing such request, one begins to observe explainable AI (XAI) models requested by medical services, as clinicians demand clinical evidence to justify machine learning-based predictions~\cite{molnar2020interpretable, ghassemi2021false}.
By simulating clinical reasoning -- retrieving similar cases, identifying abnormalities, and providing diagnostic explanations -- one can better integrate [X]AI systems into existing clinical workflows, improve trust, and enhance decision-making~\cite{tjoa2020survey, mckinney2020international, wysocki2023assessing}.
Nevertheless, despite advances, creating AI models that reliably align predictions with interpretable evidence remains an ongoing challenge.

\subsection{Current Gaps in the Field}
Existing AI models for AD diagnosis predominantly focus on maximizing predictive accuracy while often overlooking the critical need for explainability and reliability~\cite{warren2023functional, kaur2024systematic, ebrahimighahnavieh2020deep, fathi2022early, frizzell2022artificial}.
Many models treat explainability as an afterthought, adding \textit{post hoc} explanations that may not reflect the prediction process~\cite{de2020explainable, zhao2021baylime, lundberg2020local, scott2017unified, arrieta2020explainable}.
Such disconnection, therefore, limits the adaptation of AI models in clinical practices, as the quality of \textit{post hoc} explanations often fail to meet the rigorous standards required in medical decision-making.

Moreover, the modularity of diagnostic reasoning -- where distinct steps such as abnormality identification, evidence retrieval, and final diagnosis are separated -- has not been effectively captured by existing methods~\cite{ching2018opportunities}.
As a result, the lack of modular design reduces transparency and overlooks or obfuscates diagnostic errors at specific stages of the AI pipeline.

Taken together, there is an urgent need to find a unified approach that integrates evidence-driven reasoning and modular workflows.

\subsection{Objective}
In this study, we propose \textbf{VisTA} (Vision-Text Alignment Model), a new multimodal language-vision framework designed to optimize predictability, reliability, and explainability in AD diagnosis.
In brief, VisTA aligns radiology images with verified reference abnormalities and descriptions using a language-vision model assisted by contrastive learning. More specifically, we built VisTA from BiomedCLIP, a pre-trained language-vision model for medical imaging, and then fine-tuned it using contrastive learning~\cite{zhang2024biomedclip, zhang2022contrastive}.
This evidence-driven reasoning process provides AD predictions, including (i) the type of abnormality, (ii) similarity scores with reference cases, and (iii) detailed descriptions of abnormalities.
The predictive process simulates real-world clinical diagnostic workflows and yields outcomes that achieve accuracy, reliability, and interpretability.

The model's modular architecture ensures transparency at each diagnostic stage. This allows clinicians to verify intermediate results, identify potential errors, and interpret final predictions with greater confidence.
This design aligns with clinical requirements in practice, where clinical-decision support tools must integrate seamlessly with existing workflows.

\subsection{Contributions}
The key contributions of this work are as follows:
\begin{enumerate}
    \item We build \textbf{VisTA}, a multimodal language-vision model fine-tuned using contrastive learning for radiology image analysis with evidence-based reasoning. Despite being fine-tuned on only 170 samples, VisTA outperforms pre-trained models trained on millions of samples.
    
    \item We create \textbf{MINDset} (Multimodal Imaging and NeuroDiagnostic data-set), a verified reference data, containing radiology images, abnormality types, and descriptive explanations. Others can use MINDset to evaluate a machine learning model's explainability and diagnostic reliability.
    
    \item We introduce a \textbf{modular} diagnostic machine learning framework that mirrors clinical workflows. It provides a step-wise approach that is transparent and easy to detect errors.
    
    \item We demonstrate that, using comprehensive evaluations, VisTA delivers high \textbf{prediction accuracy} and \textbf{interpretability}, with results aligned with assessments from human experts. This strengthens clinical trustworthiness of the model.
    
    \item We illustrate, via case studies, how VisTA simulates \textbf{clinical reasoning} by retrieving similar cases, assigning meaningful similarity scores, and generating reliable evidence-driven diagnosis.
\end{enumerate}

By addressing existing critical gaps in explainability and reliability, this work contributes to the ever-growing research on multimodal AI in healthcare and provides a foundation for future advancements in evidence-driven diagnostic tools.

\subsection{Statement of Significance}
\subsubsection{Problem or Issue}
Diagnosing AD using radiology images is challenging due to complex data and subjective interpretation.
Existing AI models often lack explainability and reliability.

\subsubsection{What is Already Known}
AI-based approaches, including multimodal models, are promising in AD diagnosis.
Most methods, however, prioritize prediction accuracy over explainability. Nor do they align with practical clinical workflows, reducing their effectiveness in real-world applications.

\subsubsection{What This Paper Adds}
This paper introduces an evidence-driven, explainable diagnostic framework using a multimodal language-vision model enhanced by contrastive learning.
The model predicts AD outcomes, retrieves reference abnormalities, provides similarity scores, and generates interpretable explanations.
This modular design simulates clinical reasoning, enhancing reliability and transparency.

\subsubsection{Who Would Benefit from the New Knowledge in this Paper}
Our interpretable diagnostic tool, modular framework for innovation, and reliable AI-based clinical workflows benefit AI researchers, clinicians, and healthcare organizations.

\section{Related Work}
\subsection{Alzheimer's Disease Prediction}
AI has shown a great deal of promise in Alzheimer's disease studies, particularly in the territory of automated disease prediction using neuroimaging data, such as magnetic resonance imaging (MRI) and positron emission tomography (PET) scans~\cite{jack2018nia, menagadevi2024machine, illakiya2023automatic, yao2023artificial, warren2023functional, ebrahimighahnavieh2020deep, frizzell2022artificial}. Deep learning techniques, especially convolutional neural networks (CNNs), have demonstrated high accuracy in identifying structural abnormalities associated with AD progression~\cite{menagadevi2024machine, illakiya2023automatic, abdulazeem2021cnn, farooq2017deep, khagi20203d, folego2020alzheimer, el2024novel, hu2023conv, basaia2019automated}. To achieve this, these models typically leverage large datasets to extract features indicative of cortical atrophy, hippocampal shrinkage, and white matter hyperintensities (WMH)~\cite{weiner2013alzheimer, dadar2022white}.

Despite advances, most deep learning models focus solely on predictive accuracy, and with a black-box architecture, often overlooking explainability, transparency, and modularity.
Furthermore, existing machine learning approaches seldom follow the reasoning process on which clinicians rely in practice, such as comparing findings to reference cases or providing detailed explanations for abnormalities. 
As such, it is difficult for clinicians to verify, and, therefore, to trust the machine-based predictions, especially when outcomes do not align with observed data~\cite{ghassemi2021false, arrieta2020explainable}. Thus, there is a need to address these limitations if one wants to implement AI in real-work clinical workflows.

\subsection{Multimodal AI}
The integration of multimodal data, such as combining radiology images with text-based clinical notes, is powerful in biomedical applications~\cite{soenksen2022integrated, acosta2022multimodal, alsaad2024multimodal}. For example, models like BiomedCLIP, ConVIRT, and MedCLIP leverage both visual and textual modalities to enhance the machine's understanding of complex medical scenarios~\cite{zhang2022contrastive, wang2022medclip, zhang2024biomedclip}. By aligning image and text embeddings, these models extract features containing richer information and obtain augmented contextualization. Consequently, it improves the accuracy of disease classification and anomaly detection.

Although multimodal analyses that combine images and texts have shown utility in domains such as breast cancer detection and ophthalmology, their application to neurodegenerative diseases is limited~\cite{nakach2024comprehensive, qian2024multimodal, abdullakutty2024histopathology, xu2022multi, wang2024advances, mihalache2024accuracy}. For AD, there is often a wealth of available neuroimaging data and textual descriptions, and there is, therefore, considerable potential to extend multimodal methods to AD to improve both diagnostic accuracy and interpretability.

\subsection{Contrastive Learning}
Contrastive learning has gained attention as an effective technique for representation learning, particularly in scenarios with limited labeled data~\cite{chen2020simple, he2020momentum}. Contrastive learning trains models to align embeddings of related pairs (e.g., an image and its corresponding description) while separating unrelated pairs; it enhances feature discrimination and robustness~\cite{zhang2022contrastive}. Such a learning technique has seen success in handling medical imaging tasks, such as lesion detection and segmentation, where it improves generalization by learning from diverse data distributions~\cite{azizi2021big, chaitanya2020contrastive, krishnan2022self, zang2021scehr}.

In Alzheimer's disease research, contrastive learning can be used to align radiology images with verified reference abnormalities and descriptions. By linking input images to meaningful reference cases, this alignment not only improves diagnostic performance but also facilitates explainability ~\cite{zhang2024biomedclip}. Particularly, fine-tuning pre-trained models, such as BiomedCLIP, with contrastive learning offers a promising pathway to address the challenges of limited labeled datasets and enhance reliability.

\subsection{Explainable AI}
Explainable AI techniques are critical for building trust in AI systems, especially in healthcare~\cite{holzinger2018machine, amann2020explainability, sadeghi2024review, molnar2020interpretable}. Popular methods include saliency maps, attention mechanisms, and feature attribution techniques that highlight important regions of an image or influential features in a prediction~\cite{tjoa2020survey, lundberg2020local, scott2017unified, li2023deep, borys2023explainable, wollek2023attention}. Many of these methods, however, generate \textit{post hoc} explanations that may not directly align with the model's prediction process, leading to concerns about their reliability~\cite{ghassemi2021false, arrieta2020explainable}.

For AD diagnosis, generating explanations that follow clinical reasoning, for example, retrieving similar cases, identifying abnormalities, and providing detailed descriptions to justify predictions, is particularly important. Evidence-driven reasoning, as demonstrated in our work, addresses these limitations by making explanations an integral part of the diagnostic process~\cite{mckinney2020international}. This not only enhances transparency but also aligns AI tools with the modular workflows commonly used in clinical practice.

\section{Methods}

\subsection{Datasets}
\begin{figure}[!t]
    \centering
    \includegraphics[width=\textwidth]{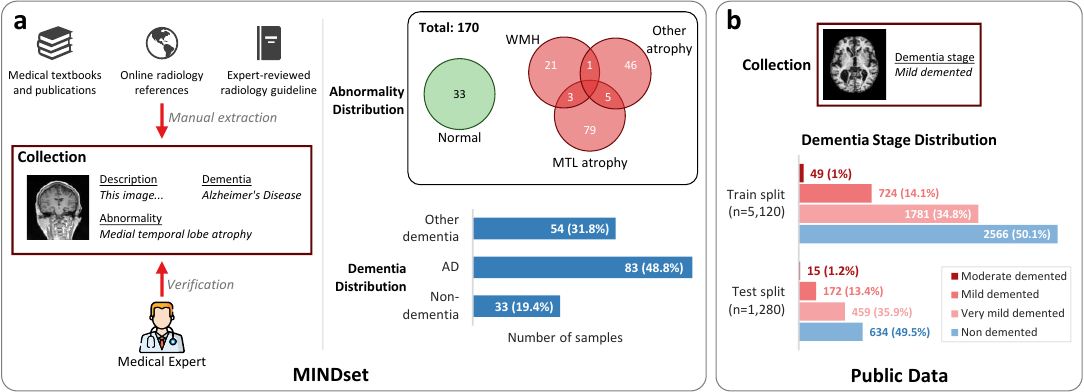}
    \caption{
        \textbf{Overview of Datasets for Abnormality Retrieval and Alzheimer's Disease Prediction.} 
        Summary of datasets used in the study, including the constructed reference dataset and publicly available Alzheimer's disease datasets. (a) The reference dataset (MINDset) contains medical radiology images paired with descriptions and abnormality types (e.g., normal, medial temporal lobe atrophy, white matter hyperintensities, and other atrophy), verified by medical experts. (b) The public dataset includes MRI scans and diagnostic labels for Alzheimer's disease prediction. Key statistics, such as the number of images, types of abnormalities, and verification process, are also shown.
    }
    \label{fig:datasets}
\end{figure}

This study uses two datasets: a reference dataset constructed for abnormality retrieval and verification and a publicly available AD prediction dataset.
The constructed reference dataset pairs radiology images with textual descriptions and abnormality types, forming the backbone of the explainability and reasoning process.
The public dataset, focused on MRI-based AD classification, provides the basis for testing prediction accuracy.
A summary of these datasets, including key statistics, is illustrated in Figure~\ref{fig:datasets}.

\subsubsection{Collection and Verification of the MINDset Data}
We constructed the MINDset (Multimodal Imaging and NeuroDiagnostic dataset), the reference dataset for this study, to support abnormality retrieval and explainability tasks. Each entry in the dataset consists of a 2D radiology image paired with a textual description and an associated abnormality type. The abnormality types are categorized into four types:
\begin{itemize}
    \item \texttt{`normal'} for Normal.
    \item \texttt{`mtl\_atrophy'} for Medial Temporal Lobe (MTL) Atrophy.
    \item \texttt{`wmh'} for White Matter Hyperintensities (WMH).
    \item \texttt{`other\_atrophy'} for Other Atrophy.
\end{itemize}

To curate the dataset, we extracted radiology images and corresponding descriptions from medical textbooks, publications~\cite{mindset-ref-01, mindset-ref-02, mindset-ref-03, mindset-ref-04}, online radiology references~\cite{radiopaedia}\footnote{\url{http://radiopaedia.org/}}, and expert-reviewed radiology guideline~\cite{radiologyassistant}\footnote{\url{https://radiologyassistant.nl/neuroradiology/dementia/role-of-mri}}. We selected the images carefully to represent a diverse range of abnormality presentations; this ensures comprehensive coverage of the four categories. The textual descriptions provide detailed explanations of each abnormality; this supports interpretability and promotes clinical reasoning processes.

To ensure the clinical accuracy and relevance of the dataset, all entries underwent a rigorous verification process conducted by a medical expert specialized in neuroimaging. This process included cross-referencing descriptions with established diagnostic guidelines and validating abnormality classifications based on radiological features.

The final dataset includes $170$ MRI brain images, each accompanied by detailed clinical information, including textual descriptions, type of abnormalities, and the type of dementia, with a particular focus on MTL atrophy due to its distinctive nature in Alzheimer's disease. This carefully curated and verified dataset serves as the backbone of the proposed model's evidence-driven reasoning framework, enabling reliable and explainable predictions.

\subsubsection{The Public Dataset for AD Prediction}
For disease prediction, this study utilizes the publicly available Alzheimer MRI Disease Classification Dataset~\cite{alzheimer_mri_dataset} from Hugging Face~\footnote{\url{https://huggingface.co/datasets/Falah/Alzheimer_MRI}}.
This dataset provides 2D brain MRI slices labeled into four diagnostic categories:
\begin{itemize}
    \item \texttt{`0'} for Non-Demented.
    \item \texttt{`1'} for Very Mild Demented.
    \item \texttt{`2'} for Mild Demented.
    \item \texttt{`3'} for Moderate Demented.
\end{itemize}

The dataset comprises a train split with $5,120$ examples and a test split with $1,280$ examples. This provides a distribution of images that covers a broad range of cases for training and evaluation. Each MRI slice offers insight into structural changes in the brain, such as hippocampal atrophy and white matter hyperintensities, which are indicative of AD progression \cite{jack2018nia, weiner2013alzheimer}. The dataset simplifies computational requirements by focusing on 2D slices while retaining key diagnostic features.

The use of this dataset ensures consistency with prior studies and provides a benchmark for evaluating the prediction performance of the proposed VisTA model. To ensure uniformity in the dementia labeling, we reclassify the dataset into two categories:
\begin{itemize}
    \item \texttt{`0'} for Non-demented.
    \item \texttt{`1'} for Demented.
\end{itemize}

\subsection{VisTA Model Architecture}  

\begin{figure}[!t]
    \centering
    \includegraphics[width=\textwidth]{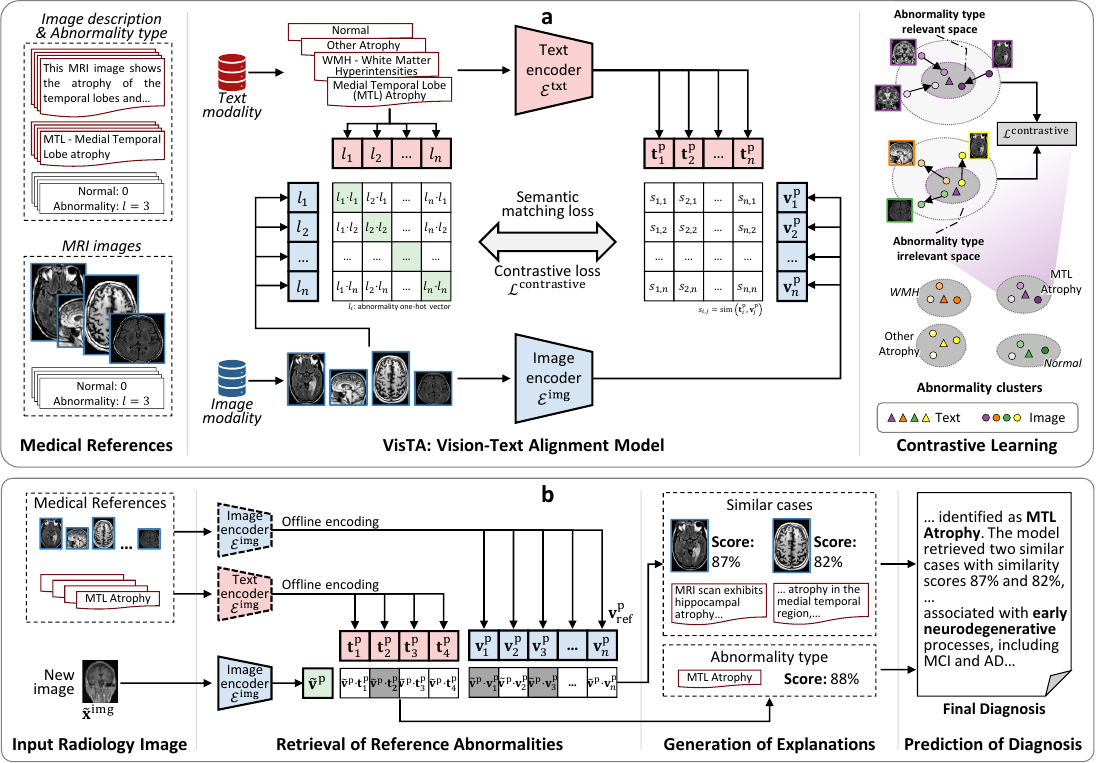}
    \caption{
        \textbf{Model Architecture and Workflow of the Proposed Method.} 
        \textbf{\textit{(a)}} The schematic representation of the VisTA architecture, showing its pre-trained components and the fine-tuning process using contrastive learning to align image and text embeddings. The architecture incorporates multimodal learning to capture relationships between radiology images and their descriptions and abnormality type.
        \textbf{\textit{(b)}} The workflow of the VisTA framework, with its end-to-end pipeline: (1) input radiology image, (2) retrieval of reference abnormalities with similarity scoring, (3) generation of descriptive explanations, and (4) prediction of Alzheimer's disease based on the retrieved evidence. The pipeline highlights the modularity and explainability of VisTA: it ensures reliability and alignment with real-world clinical diagnostic workflows.
    }
    \label{fig:architecture}
\end{figure}

The proposed VisTA model is based on the BiomedCLIP architecture~\cite{zhang2024biomedclip}, a multimodal language-vision framework, and further fine-tuned using contrastive learning to improve medical images and text alignment.
BiomedCLIP employs a dual-encoder system consisting of a vision encoder for radiology images and a text encoder for textual descriptions. The dual encoder then projects respective image and text inputs into a shared embedding space for multimodal alignment.
We present the VisTA architecture in Figure~\ref{fig:architecture}-a.  

\subsubsection{The Vision-Text Dual Encoder}  
The vision encoder, $\mathcal{E}^{\text{img}}$, processes radiology images $\mathbf{x}^{\text{img}}$ into feature representations $\mathbf{v} \in \mathbb{R}^D$, which are further mapped into the shared embedding space using a projection head $f^\text{v}$:  
\begin{equation}
    \mathbf{v} = \mathcal{E}^{\text{img}}(\mathbf{x}^{\text{img}}), \quad \mathbf{v}^\text{p} = f^\text{v}(\mathbf{v}).
\end{equation}
The text encoder, $\mathcal{E}^{\text{txt}}$, takes textual descriptions $\mathbf{x}^{\text{txt}}$ as input and outputs feature representations $\mathbf{t} \in \mathbb{R}^M$, which are similarly projected into the shared embedding space:  
\begin{equation}
    \mathbf{t} = \mathcal{E}^{\text{txt}}(\mathbf{x}^{\text{txt}}), \quad \mathbf{t}^\text{p} = f^\text{t}(\mathbf{t}).
\end{equation}
Here, $\mathbf{v}^\text{p}$ and $\mathbf{t}^\text{p}$ share the same dimensionality, $P$, enabling effective contrastive alignment.  

\subsubsection{Adaptations for Abnormality Retrieval}
To address the limited size of the training dataset and enhance explainability, in VisTA, we made several adaptations to the BiomedCLIP framework:  

\begin{itemize}
    \item \textbf{Fixed abnormality type text for text modality}:
    Instead of using `dynamic' text description for each image, VisTA used fixed abnormality type text (i.e., ``Normal,'' ``Medial Temporal Lobe Atrophy,'' ``White Matter Hyperintensities,'' and ``Other Atrophy'') to represent each category.
    VisTA aligned all images within the same abnormality category with their corresponding fixed abnormality type text during contrastive learning.
    This ensures that images in the same category cluster closely in the shared embedding space, while images from different categories are separated.
    
    \item \textbf{Small dataset optimization}:
    VisTA fine-tuned the BiomedCLIP using the constructed reference dataset, which is relatively small but highly curated. 
    VisTA adapted contrastive learning to emphasize intra-category alignment and inter-category separation, using as much information of the limited data as possible.
    
    \item \textbf{Category-based regularization}:
    VisTA added a regularization term to further strengthen the alignment between images and their abnormality types.
    This helps stabilize training and reduces the risk of overfitting on small datasets.
\end{itemize}

\subsubsection{Contrastive Loss with Adaptations}  
The contrastive loss used in this study aligns image and text embeddings while separating mismatched pairs:

\begin{equation}
    \mathcal{L}^{\text{contrastive}} = - \frac{1}{N} \sum_{i=1}^N \log \frac{\exp\left( \text{sim}\left( \mathbf{v}^\text{p}_i, \mathbf{t}^\text{p}_i \right) / \tau \right)}{\sum_{j=1}^N \exp\left( \text{sim}\left( \mathbf{v}^\text{p}_i, \mathbf{t}^\text{p}_j \right) / \tau \right)}.
\end{equation}
Here, $\text{sim}\left( \mathbf{v}^\text{p}_i, \mathbf{t}^\text{p}_j \right)$ denotes the cosine similarity between the image embedding $\mathbf{v}^\text{p}_i$ and the text embedding $\mathbf{t}^\text{p}_j$, and $\tau$ is a temperature parameter. By using fixed abnormality type text for $\mathbf{t}^\text{p}_j$, the model ensures that images are well clustered based on their abnormality categories.  

\subsubsection{An Overview of VisTA Workflow}  
VisTA processed radiology images through the vision encoder and passed the fixed textual descriptions, representing abnormality categories, through the text encoder. It then aligned the embeddings from both modalities through contrastive learning, creating a shared space for evidence retrieval. This adaptation enhances explainability and facilitates robust abnormality classification and AD prediction tasks.

\subsection{Explainability and Reasoning Process}  

VisTA integrates abnormality identification, similarity retrieval, and descriptive explanation generation as core components of the diagnostic workflow (Figure~\ref{fig:architecture}-b).
These components are modular and reflect the real-world clinical diagnostic process, where identifying and reasoning about abnormalities is an essential precursor to disease classification.  

\subsubsection{Abnormality Identification and Similarity Retrieval}  
Technically, VisTA processed radiology images, $\tilde{\mathbf{x}}^{\text{img}}$, through the vision encoder, $\mathcal{E}^{\text{img}}$, to generate embeddings, $\tilde{\mathbf{v}}^\text{p}$. It utilized the embeddings in two key comparison steps, explained below.

\paragraph{Comparison to Textual Embeddings}
VisTA compared the vision embedding $\tilde{\mathbf{v}}^\text{p}$ to fixed textual embeddings, $\mathbf{t}^\text{p}$, corresponding to predefined abnormality categories (e.g., ``Normal,'' ``Medial Temporal Lobe Atrophy'').
It used the cosine similarity to calculate the alignment:  
\begin{equation}
   \text{sim} \left( \tilde{\mathbf{v}}^\text{p}, \mathbf{t}^\text{p} \right) = \frac{\tilde{\mathbf{v}}^\text{p} \cdot \mathbf{t}^\text{p}}{\|\tilde{\mathbf{v}}^\text{p}\| \|\mathbf{t}^\text{p}\|}.
\end{equation}
It then used the similarity scores to assign the input image to the most probable abnormality type:  
\begin{equation}
   \hat{y}^{\text{abn}} = \argmax_{\mathbf{t}^\text{p}} \left(\text{sim} \left( \tilde{\mathbf{v}}^\text{p}, \mathbf{t}^\text{p} \right) \right),
\end{equation}
where $\hat{y}^{\text{abn}}$ represents the predicted abnormality type.  

\paragraph{Comparison to Reference Image Embeddings}
VisTA also compared the vision embedding $\tilde{\mathbf{v}}^\text{p}$ to the embeddings of all reference images in the constructed dataset.
For each reference image embedding, it computed the cosine similarity, $\mathbf{v}^\text{p}_{\text{ref}}$, as:  
\begin{equation}
   \text{sim} \left( \tilde{\mathbf{v}}^\text{p}, \mathbf{v}^\text{p}_{\text{ref}} \right) = \frac{\tilde{\mathbf{v}}^\text{p} \cdot \mathbf{v}^\text{p}_{\text{ref}}}{\|\tilde{\mathbf{v}}^\text{p}\| \|\mathbf{v}^\text{p}_{\text{ref}}\|}.
\end{equation}  
This comparison retrieves a set of reference images that are most similar to the input image. These retrieved cases provide evidence supporting the abnormality identification and inform the reasoning process.  

This dual comparison approach ensures that VisTA provides both categorical predictions and concrete evidence in the form of similar cases, enhancing explainability and reliability.  

\subsubsection{Description Generation}  
The explanation process combines multiple sources of evidence to generate a descriptive summary of the input image:  
\begin{itemize}
    \item \textbf{Predicted abnormality type}:
    VisTA included pre-defined descriptions associated with the predicted abnormality type, $\hat{y}^{\text{abn}}$, to highlight general characteristics of the identified abnormality.
    
    \item \textbf{Similar reference cases}:
    VisTA used the retrieved reference images and their similarity scores to provide additional context. For each reference case, it incorporated the textual description and similarity score into the explanation using a template-based approach.
    
    \item \textbf{Lightweight LLM refinement}: VisTA refined the final explanation using a lightweight language model, which consolidates the descriptive elements and ensures clarity, fluency, and clinical relevance.
\end{itemize}

The resulting description offers a comprehensive explanation of the input image, integrating both categorical and case-specific evidence. This enhances interpretability and aligns with real-world clinical reasoning workflows.  

\subsection{Final Outcome Prediction}  

The final stage of VisTA integrates evidence from the abnormality identification and reasoning process to predict Alzheimer's disease. This modular design mirrors clinical workflows in practice, where abnormalities identified in radiology images inform subsequent diagnostic decisions.  

\subsubsection{Evidence Integration for AD Prediction}  
The AD prediction task incorporates information from the earlier stages of the pipeline.
Specifically, VisTA combined the vision embedding, $\tilde{\mathbf{v}}^\text{p}$, with the textual embedding of the predicted abnormality type, $\mathbf{t}^\text{p}_{\hat{y}^{\text{abn}}}$, and the embeddings of retrieved reference cases, $\mathbf{v}^\text{p}_{\text{ref}}$.
A classification head, $f^{\text{AD}}$, takes these inputs and predicts the AD outcome:
\begin{equation}
    \hat{y}^{\text{AD}} = f^{\text{AD}}\left(\tilde{\mathbf{v}}^\text{p}, \mathbf{t}^\text{p}_{\hat{y}^{\text{abn}}}, \mathbf{v}^\text{p}_{\text{ref}}\right).
\end{equation}
This multi-source integration ensures AD prediction is directly informed by both categorical abnormality classification and evidence from similar cases.  

\subsubsection{Clinical Alignment and Modularity}  
VisTA's modular design allows clinicians to independently evaluate the intermediate outputs of the modeling flow, such as abnormality type predictions, similarity scores, and retrieved reference cases.
Such transparency enables error diagnosis and iterative model refinement. For example, discrepancies between the retrieved evidence and the final AD prediction can be traced to specific earlier steps and improved, allowing continuous model improvement.  

By integrating evidence-driven reasoning into a modular framework, VisTA enhances both reliability and clinical applicability, ensuring that machine-based predictions align with real-world diagnostic practices.  

\subsection{Evaluation Metrics}  

To comprehensively assess the performance and reliability of VisTA, we employed a combination of quantitative and qualitative evaluation metrics. These metrics evaluate both the predictive accuracy and the explainability of the model, ensuring alignment with real-world clinical expectations.  

\subsubsection{Predictive Performance Metrics}  
We used the following metrics to quantify VisTA's performance on abnormality classification and AD prediction tasks.

\textit{Accuracy}
measures the proportion of correctly classified samples, providing an overall indication of model performance.  
\begin{equation}
    \text{Accuracy} = \frac{\text{TP} + \text{TN}}{\text{TP} + \text{TN} + \text{FP} + \text{FN}}.
\end{equation}  
Here,
True Positives (TP) represent cases where the model correctly identifies a positive class (e.g., an abnormality or AD case);
True Negatives (TN) are cases where the model correctly identifies a negative class (e.g., a normal or non-AD case);
False Positives (FP) occur when the model incorrectly classifies a negative case as positive;
and False Negatives (FN) occur when the model fails to classify a positive case.  

\textit{Area Under the ROC Curve (AUC-ROC)} evaluates the model's ability to distinguish between classes: higher AUC-ROC values indicating better results discrimination results.

\textit{Precision} indicates how accurate the model's positive predictions are (e.g., an abnormality or AD case).  
\begin{equation}
    \text{Precision} = \frac{\text{TP}}{\text{TP} + \text{FP}}.
\end{equation}

\textit{Sensitivity (Recall)} indicates the model's ability to correctly identify positive cases (e.g., an abnormality or AD case).  
\begin{equation}
    \text{Sensitivity} = \frac{\text{TP}}{\text{TP} + \text{FN}}.
\end{equation}  

\textit{Specificity} reflects the model's ability to correctly identify negative cases (e.g., a normal or non-AD case).  
\begin{equation}
    \text{Specificity} = \frac{\text{TN}}{\text{TN} + \text{FP}}.
\end{equation}  

\textit{F1-Score} provides a harmonic mean of precision and sensitivity, offering a balanced measure when classes are imbalanced:
\begin{equation}
    \text{F1-Score} = 2 \cdot \frac{\text{Precision} \cdot \text{Recall}}{\text{Precision} + \text{Recall}}.
\end{equation}

These metrics ensure that the model is evaluated on its overall correctness and its ability to minimize false positives and false negatives, which is critical in clinical applications.  

\subsubsection{Human-Expert Validation}  
To assess the reliability and clinical relevance of the model's predictions and explanations, a human-expert validation process was conducted:  

\begin{enumerate}
    \item \textbf{Abnormality identification}:
    Expert radiologists reviewed the predicted abnormality types and retrieved reference cases. Agreement rates between the model and experts were calculated to quantify the accuracy and reliability of the abnormality classification.
    
    \item \textbf{Explanation quality}:
    Medical experts qualitatively evaluated the generated explanations, including retrieved reference cases, similarity scores, and descriptive summaries. Criteria for evaluation included clinical relevance, interpretability, and consistency with standard diagnostic guidelines.
    
    \item \textbf{Dementia diagnosis}:
    Final dementia predictions were compared with expert diagnoses to measure agreement and highlight areas for model improvement.
\end{enumerate}

This combination of automated metrics and human validation ensures that the model's performance is not only quantitatively robust but also clinically meaningful. By prioritizing both predictive accuracy and explainability, the proposed method aligns with the requirements of real-world diagnostic workflows.  

\section{Results}

\subsection{Experimental Setup}
To evaluate the efficacy of VisTA, we conducted experiments on the constructed MINDset and a publicly available dataset~\cite{alzheimer_mri_dataset} for AD classification.
The experimental setup ensures rigorous validation while maintaining separation between training and test data to prevent data leakage.

\subsubsection{Data Setup}
\paragraph{MINDset}
The MINDset consists of $170$ samples, which were split into a training set of $120$ samples and a test set of $50$ samples. The test set remained entirely separate throughout model training to ensure an unbiased evaluation of performance.

\paragraph{Public Dataset}
For the public dataset, we utilized the designated test split, which includes $1,280$ MRI images with dementia labels categorized into different severity levels. Since this dataset does not contain abnormality annotations, we used it solely for evaluating the AD prediction task.  

\subsubsection{Evaluation Setup}
For both datasets, the experimental workflow consists of three sequential steps:  
\begin{enumerate}
    \item Inputting an MRI image into the model,
    
    \item Extracting the predicted abnormality type from the image,
    
    \item Computing the probability of Alzheimer's disease/Dementia based on the identified abnormality.
\end{enumerate}

The evaluation includes two primary tasks:
\textbf{Abnormality retrieval} assesses the model's ability to identify abnormalities present in the MRI images correctly. We summarize the performance on this task in Table~\ref{tab:comparative_analysis}-a.
\textbf{Disease prediction} evaluates the model's ability to predict dementia based on the identified abnormality. This step involves computing the conditional probability of each dementia type/stage given the predicted abnormality. We report the results in Table~\ref{tab:comparative_analysis}-b and~\ref{tab:comparative_analysis}-c.

Since \textbf{MINDset} contains both abnormality types and dementia labels, we report performance for both \textit{Abnormality Retrieval} and \textit{Disease Prediction} tasks.
Given that the \textbf{public dataset} includes only dementia labels, we evaluate and report performance solely for \textit{Disease Prediction}, where abnormalities are inferred before estimating dementia probabilities.

\subsection{Model Performance}

Table~\ref{tab:comparative_analysis} compares VisTA with the baseline models BiomedCLIP~\cite{zhang2024biomedclip}, MedCLIP~\cite{wang2022medclip}, and ConVIRT~\cite{zhang2022contrastive} across various configurations.
Although using only $170$ samples for fine-tuning,
VisTA demonstrated significant improvements in all tasks, particularly in abnormality retrieval and dementia prediction.
Aligning vision and text embeddings via fine-tuning yielded a substantial boost of accuracy and explainability. This showed the effectiveness of its evidence-driven reasoning process.

\begin{table}[!t]
\centering
\caption{\textbf{Model Comparisons regarding Abnormality Retrieval and Disease Prediction Tasks.}
}
\label{tab:comparative_analysis}
\resizebox{\textwidth}{!}{%
\begin{tabular}{lcccccc}
\toprule
\multicolumn{1}{c}{\textbf{Method}} & \textbf{\begin{tabular}[c]{@{}c@{}}Accuracy\\ (\%)\end{tabular}} & \textbf{AUC-ROC} & \textbf{\begin{tabular}[c]{@{}c@{}}Sensitivity\\ (\%)\end{tabular}} & \textbf{\begin{tabular}[c]{@{}c@{}}Specificity\\ (\%)\end{tabular}} & \textbf{\begin{tabular}[c]{@{}c@{}}Precision\\ (\%)\end{tabular}} & \textbf{\begin{tabular}[c]{@{}c@{}}F1-score\\ (\%)\end{tabular}} \\ \hline
\multicolumn{7}{l}{\cellcolor[HTML]{C0C0C0}\textit{\textbf{a. Abnormality Retrieval Performance$^\dagger$}}} \\ \hline
\multicolumn{7}{l}{\textit{Performance on MINDset}} \\ \hline
BiomedCLIP & 26.00 & 0.74 & 37.85 & 80.04 & 25.96 & 27.39 \\
MedCLIP & 24.00 & 0.52 & 22.96 & 75.51 & 36.14 & 18.23 \\
ConVIRT & 26.00 & 0.56 & 25.00 & 75.00 & 6.50 & 10.32 \\
\textbf{VisTA}$^\S$ &  &  &  &  &  &  \\
\quad- Description & 48.00 & 0.76 & 50.19 & 82.34 & 53.58 & 45.06 \\
\quad- Abnormality & \textbf{74.00} & \textbf{0.87} & \textbf{72.30} & \textbf{90.53} & \textbf{74.02} & \textbf{72.05} \\
\quad- Summary & 70.00 & 0.85 & 66.96 & 89.15 & 72.78 & 68.35 \\
\quad- All & 46.00 & 0.73 & 46.77 & 81.25 & 43.91 & 40.86 \\ \hline
\multicolumn{7}{l}{\cellcolor[HTML]{C0C0C0}\textit{\textbf{b. Dementia Prediction Performance$^\ddagger$}}} \\ \hline
\multicolumn{7}{l}{\textit{Performance on MINDset}} \\ \hline
BiomedCLIP & 30.00 & 0.57 & 17.50 & 80.00 & 77.77 & 28.57 \\
MedCLIP & 58.00 & 0.56 & 65.00 & 30.00 & 78.78 & 71.23 \\
ConVIRT & 80.00 & 0.62 & 100.00 & 0.00 & 80.00 & 88.88 \\
\textbf{VisTA}$^\S$ &  &  &  &  &  &  \\
\quad- Description & 84.00 & 0.63 & \textbf{97.50} & 30.00 & 84.78 & 90.70 \\
\quad- Abnormality & \textbf{88.00} & \textbf{0.82} & 95.00 & \textbf{60.00} & \textbf{90.48} & 92.68 \\
\quad- Summary & \textbf{88.00} & 0.77 & \textbf{97.50} & 50.00 & 88.63 & \textbf{92.86} \\
\quad- All & 78.00 & 0.53 & 92.50 & 20.00 & 82.22 & 87.06 \\ \hline
\multicolumn{7}{l}{\textit{Performance on Public Data}} \\ \hline
BiomedCLIP & 49.45 & 0.50 & 0.15 & \textbf{99.68} & \textbf{100.00} & 0.31 \\
MedCLIP & 50.70 & 0.51 & 43.81 & 57.73 & 51.36 & 47.28 \\
ConVIRT & 50.47 & 0.53 & 100.00 & 0.00 & 50.46 & 67.08 \\
\textbf{VisTA}$^\S$ & \multicolumn{1}{l}{} & \multicolumn{1}{l}{} & \multicolumn{1}{l}{} &  &  &  \\
\quad- Description & 56.80 & 0.59 & 66.10 & 43.84 & 55.89 & 60.56 \\
\quad- Abnormality & 51.64 & 0.54 & \textbf{100.00} & 2.40 & 51.11 & 67.64 \\
\quad- Summary & \textbf{58.28} & \textbf{0.64} & 96.13 & 19.24 & 54.96 & \textbf{69.93} \\
\quad- All & 54.61 & 0.56 & 89.16 & 17.67 & 53.14 & 66.59 \\
\hline
\multicolumn{7}{l}{\cellcolor[HTML]{C0C0C0}\textit{\textbf{c. AD Prediction Performance$^\ddagger$}}} \\ \hline
\multicolumn{7}{l}{\textit{Performance on MINDset}} \\ \hline
BiomedCLIP & 48.00 & 0.54 & 0.00 & \textbf{100.00} & 0.00 & 0.00 \\
MedCLIP & 48.00 & 0.47 & 0.00 & \textbf{100.00} & 0.00 & 0.00 \\
ConVIRT & 48.00 & 0.50 & 0.00 & \textbf{100.00} & 0.00 & 0.00 \\
\textbf{VisTA}$^\S$ &  &  &  &  &  &  \\
\quad- Description & 44.00 & 0.47 & 26.92 & 62.50 & 43.75 & 33.33 \\
\quad- Abnormality & \textbf{60.00} & \textbf{0.60} & \textbf{46.15} & 75.00 & \textbf{66.66} & \textbf{54.55} \\
\quad- Summary & 54.00 & 0.59 & \textbf{46.15} & 62.50 & 57.14 & 51.06 \\
\quad- All & 48.00 & 0.50 & 34.62 & 62.50 & 50.00 & 40.91 \\ \bottomrule
\multicolumn{7}{r}{\footnotesize\begin{tabular}[c]{@{}r@{}}
The best results in each column and experiment are in \textbf{bold} fonts.\\ 
$^\dagger$:Macro-averaged results over four (04) abnormality types.
$^\ddagger$:Binary classification results.\\
$^\S$\textbf{\textit{VisTA text modality}}:
Description: using image description;
Abnormality: using abnormality type; \\
Summary: using abnormality type and dementia type;
All: using a combination of summary and description.
\end{tabular}}
\end{tabular}%
}
\end{table}

\subsubsection{Abnormality Retrieval Performance}
On MINDset, VisTA significantly outperformed BiomedCLIP in abnormality retrieval, achieving a macro-averaged accuracy of $74\%$ and an AUC of $0.87$ when using abnormality type as the text modality. 
It outperforms BiomedCLIP (accuracy: $26\%$, AUC: $0.74$), MedCLIP (accuracy: $24\%$, AUC: $0.52$), and ConVIRT (accuracy: $26\%$, AUC: $0.56$). 
Additionally, VisTA achieved superior sensitivity ($72.3\%$) and specificity ($90.53\%$).
This suggests VisTA's capability to better distinguish between different abnormality types.
Configurations using ``abnormality type'' and ``summary'' text modalities also showed significant gains over all baselines. This highlights VisTA's robustness across different input formats.

Taken together, our results emphasize VisTA's strength in leveraging its fine-tuned alignment between images and predefined abnormality types to improve retrieval accuracy.

\subsubsection{Dementia Prediction}
For dementia prediction, VisTA demonstrated superior performance on MINDset. 
Using ``abnormality type'' as the text modality, VisTA achieved an accuracy of $88\%$, an AUC of $0.82$, and an F1-score of $92.68\%$, compared to BiomedCLIP (accuracy: $30\%$, AUC: $0.57$, F1-score: $28.57\%$), MedCLIP (accuracy: $58\%$, AUC: $0.56$, F1-score: $71.23\%$), and ConVIRT (accuracy: $80\%$, AUC: $0.62$, F1-score: $88.88\%$).
Notably, the sensitivity metrics for VisTA and BiomedCLIP were $95\%$ and $17.5\%$, respectively. This suggests VisTA's ability to identify dementia cases reliably.

VisTA also showed improved performance over other baselines on the public dataset.
While BiomedCLIP struggled to classify dementia cases, marking nearly all as non-dementia, VisTA demonstrated its ability to classify both dementia and non-dementia cases effectively.
For instance, in the ``summary'' configuration, VisTA achieved an accuracy of $58.28\%$, an AUC of $0.64$, and an F1-score of $69.93\%$, significantly outperforming BiomedCLIP's respective scores of $49.45\%$, $0.50$, and $0.31\%$.
The slightly lower performance on the public dataset highlights a need for further adaptation of VisTA to handle differences in data quality and format.

\subsubsection{AD Prediction}
For AD prediction, all baselines failed to correctly classify any AD cases and classified all as non-AD, resulting in an accuracy of $48\%$ and AUC below $54\%$, driven solely by their specificity of $100\%$ and sensitivity of $0\%$.
VisTA showed noticeable improvements. 
For example, VisTA's best configuration (using ``abnormality type'') achieved an accuracy of $60\%$, an AUC of $0.60$, and an F1-score of $54.55\%$, showing clear improvements in its ability to identify AD cases.

While VisTA's performance in AD diagnosis was better than that of other baselines, it did not reach the levels observed for dementia diagnosis or abnormality retrieval.
This discrepancy is consistent with the error analysis (see Section~\ref{sec:error_analysis}).
A significant factor limiting performance was the misclassification of AD as other types of dementia, particularly in cases where subtle or overlapping abnormalities were present such as medial temporal lobe atrophy (MTL atrophy) or white matter hyperintensities (WMH).

\subsection{Confusion Matrix Analysis}

\begin{figure}[!ht]
    \centering
    \includegraphics[width=\textwidth]{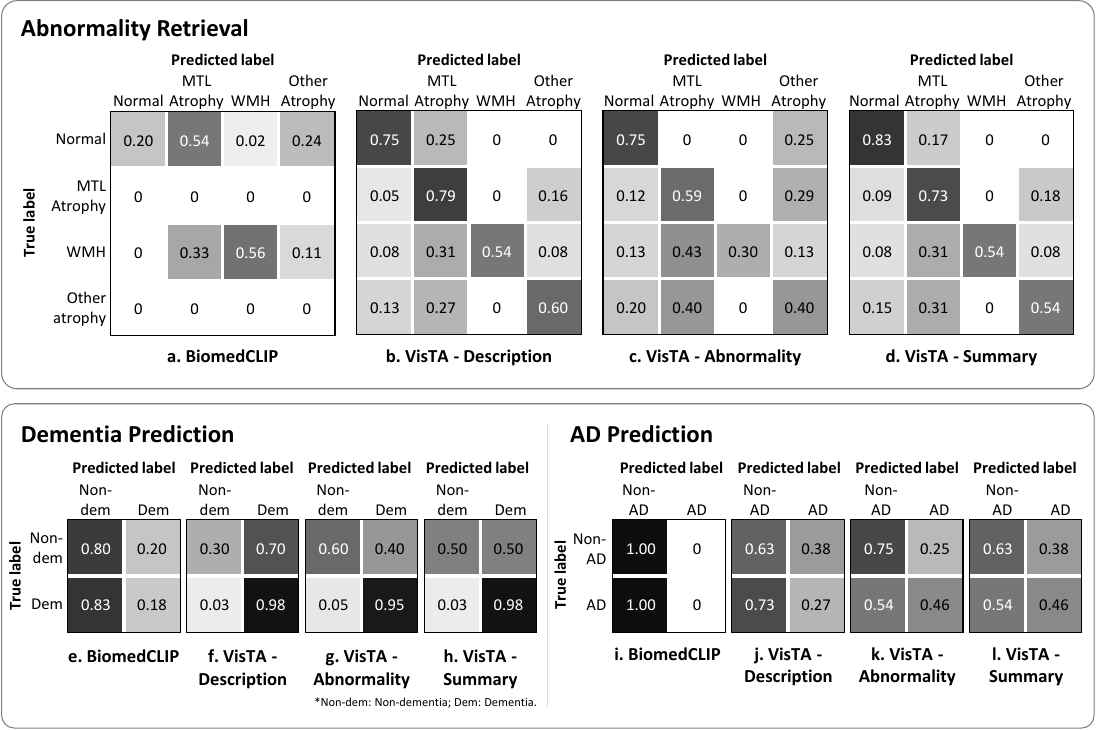}
    \caption{
        \textbf{Confusion Matrices for Abnormality Type Classification and Disease Prediction.}
        Confusion matrices illustrate the performance of pre-trained BiomedCLIP and VisTA models across abnormality type classification (Panels a-d), Dementia prediction (Panels e-h), and AD prediction (Panels i-l). Panels a-d depict the classification accuracy for identifying specific abnormality types using the pre-trained BiomedCLIP model and three versions of VisTA. Panels e-h and i-l show the performance for dementia and AD (binary) prediction. The confusion matrices highlight the improvements in classification accuracy and reduction of misclassifications achieved through contrastive learning on MINDset medical references.
    }
    \label{fig:confusion}
\end{figure}

To gain deeper insights into VisTA's performance compared to BiomedCLIP, Figure~\ref{fig:confusion} presents confusion matrices for abnormality retrieval, dementia diagnosis, and AD prediction tasks.

\paragraph{Abnormality retrieval}  
BiomedCLIP struggled to differentiate between abnormality types, with significant misclassifications. For instance, it only correctly identified $20\%$ of normal cases and misclassified $54\%$ as MTL atrophy. Similarly, it did not correctly identify any of MTL atrophy or other atrophy. This indicates its severe limitation in capturing abnormality-specific patterns.
In contrast, VisTA achieved substantial improvements, correctly identifying $83\%$ of normal cases, $73\%$ of MTL atrophy cases, $54\%$ of WMH cases, and $54\%$ of other atrophy cases under ``Summary'' configuration.
These results demonstrate VisTA's ability to better align input images with predefined abnormality types, reducing errors and providing more reliable retrieval.

\paragraph{Dementia prediction}  
BiomedCLIP exhibited poor sensitivity in dementia prediction: it classified $83\%$ of cases as non-dementia.
This suggests the model may not be able to discovery dementia cases. 
VisTA, however, showed a significant improvement in this task, achieving near-perfect sensitivity for dementia cases ($98\%$) under ``Summary'' and ``Description'' configurations. While the specificity decreased in some configurations, which we aim to improve in future works, its ability to correctly identify dementia cases suggests a critical advantage in clinical applications.

\paragraph{Alzheimer's disease prediction}  
BiomedCLIP failed to classify any AD cases correctly, labeling all inputs as non-AD.
VisTA demonstrated marked improvements, particularly in the ``Abnormality'' and ``Summary'' configurations. Both achieved higher specificity and maintained balanced sensitivity. The AD diagnosis task, however, remains challenging, with notable misclassifications between AD and non-AD cases; we aim to improve this in future works.

\subsection{Case Studies}

\begin{figure}[!t]
    \centering
    \includegraphics[width=\linewidth]{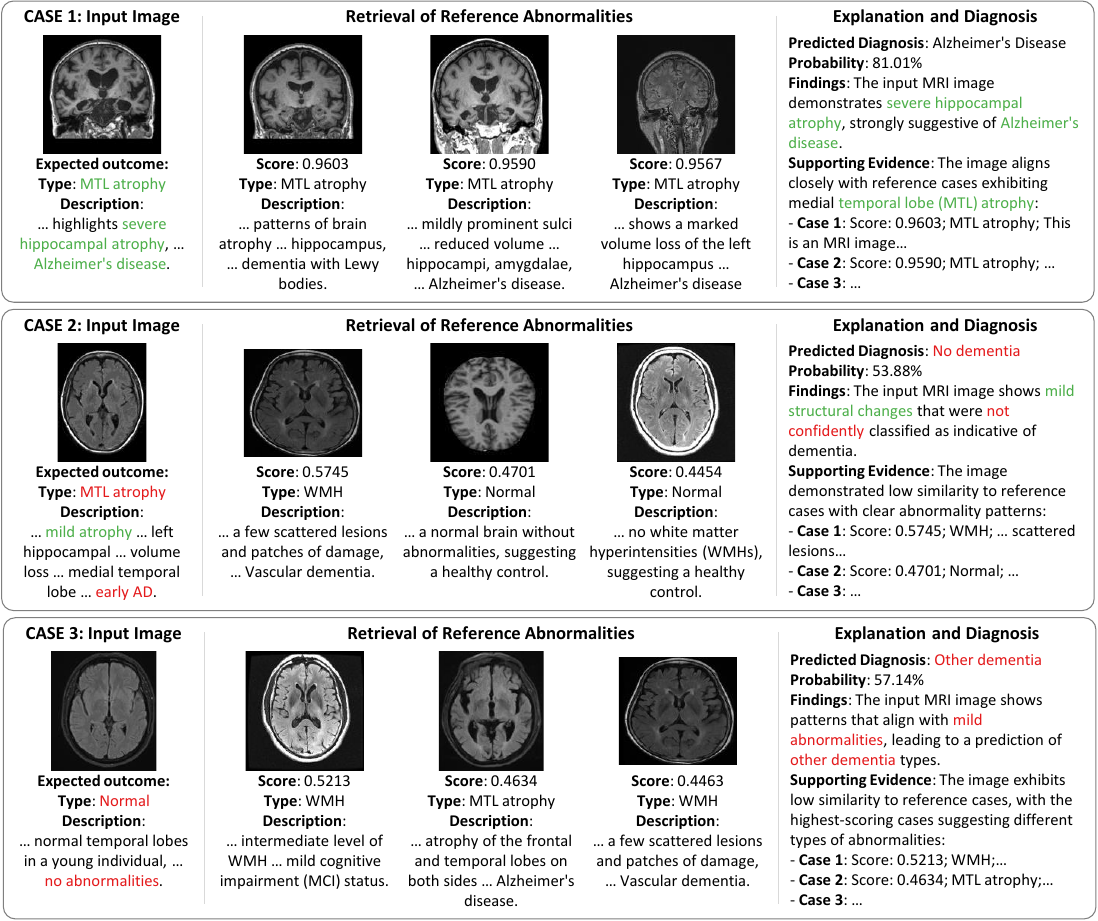}
    \caption{
        \textbf{Case Study Examples of VisTA for Delivering Evidence-Driven Explainable Alzheimer's Disease Diagnosis.} 
        VisTA input radiology images, and output explainable, evidence-based predictions that aligned with real-world clinical reasoning processes.
    }
    \label{fig:casestudy}
\end{figure}

We present three distinctive case studies to demonstrate the VisTA's explainability and alignment with real-world clinical reasoning. These examples provide a comprehensive view of VisTA's performance in abnormality retrieval, dementia-type prediction, and evidence-driven explanations.
Figure~\ref{fig:casestudy} illustrates the details of each case.

\paragraph{\textbf{CASE 1}: Accurate abnormality retrieval and disease prediction}
VisTA correctly identified the input image as exhibiting MTL atrophy, supported by a high similarity to reference images ($0.9603$, $0.9590$, $0.9567$).
The retrieved references highlighted hippocampal atrophy and characteristic patterns consistent with AD. 
VisTA predicted AD with a probability of $81.01\%,$ aligning with the clinical interpretation of the input image.
This case shows VisTA's ability to retrieve relevant references and generate accurate, evidence-based predictions, particularly for clear abnormality patterns.

\paragraph{\textbf{CASE 2}: Mild atrophy misclassified as normal}
This example presents an MRI image displaying mild MTL atrophy, which VisTA misclassified as normal.
The retrieved references included two normal image (score: $0.4701$, $0.4454$) and an unrelated WMH case (score: $0.5745$).
The model predicted no dementia with a probability of $53.88\%$. This is due likely to the subtle nature of the atrophy, which closely resembled normal brain structures.
This case suggests a limitation in VisTA's sensitivity to early-stage abnormalities, suggesting the need for further refinement to better distinguish subtle patterns.

\paragraph{\textbf{CASE 3}: False positive abnormality retrieval}
In this case, a normal image was misclassified as abnormal (specifically WMH).
The retrieved reference cases showed low similarity scores (highest score: 0.5213), with references suggesting mild cognitive impairment and vascular dementia.
The model predicted other dementia with a probability of $57.14\%$, driven by noise and subtle artifacts in the image.
This example suggests there is a need to further improve VisTA's ability to handle noisy inputs and enhance the diversity of the reference dataset to reduce such misclassifications.

Together, these case studies demonstrate VisTA's ability to provide evidence-based reasoning for its predictions, and reveal areas for improvement, particularly in handling subtle abnormalities and noisy data.

\section{Discussion}
\subsection{Embedding Space and Retrieval Performance Analysis}

The embedding space and retrieval performance analyses provide critical insights into how VisTA enhances the alignment of multimodal representations, improves abnormality retrieval, and supports diagnostic accuracy.
Figures~\ref{fig:embedding} and~\ref{fig:retrieval_performance} illustrate these findings and highlight the effectiveness of VisTA's fine-tuned architecture.

\subsubsection{Embedding Space Analysis}
\begin{figure}[!t]
    \centering
    \includegraphics[width=\linewidth]{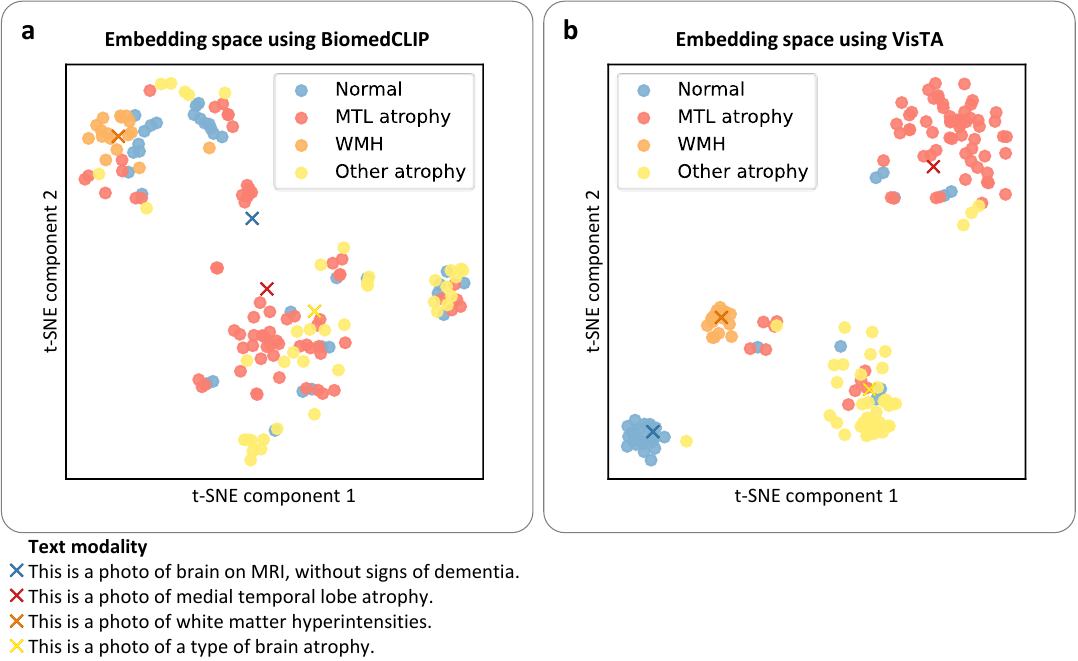}
    \caption{
        \textbf{The Embedding Space of Image and Text Representations.}
        The panels show the embedding space from the final encoder used in the diagnosis step, integrating both image embeddings and textual embeddings of abnormality types.  
        \textbf{\textit{(a)}} Embeddings from the pre-trained BiomedCLIP model, showing initial alignment between radiology image representations and fixed abnormality type text embeddings. Clustering of abnormalities is present but lacks strong separation.  
        \textbf{\textit{(b)}} VisTA embeddings, showing a clear separation between abnormality types, with improved multimodal alignment between image and text representations.  
        The results highlight the effectiveness of contrastive learning in enhancing embedding quality, ensuring better differentiation between abnormality categories.
    }
    \label{fig:embedding}
\end{figure}

Figure~\ref{fig:embedding} presents the embedding space from the final encoder used in the diagnosis step, showcasing the alignment between radiology image embeddings and textual embeddings of abnormality types.
In the pre-trained BiomedCLIP model (Figure~\ref{fig:embedding}-a), the clustering of abnormalities is evident but not well separated. This limits the model's ability to distinguish between categories effectively.

In contrast, VisTA's embeddings (Figure~\ref{fig:embedding}-b) show clearly separated clusters for each abnormality type.
The alignment between image and text representations is significantly improved, with higher intra-category cohesion and inter-category separation. These improvements stem from VisTA's adaptation of contrastive learning, which optimizes the shared embedding space to align radiology images with their corresponding abnormality type texts.
Such refined alignment enhances VisTA's ability to retrieve relevant reference cases and supports its diagnostic reasoning capabilities.

\subsubsection{Retrieval Performance Analysis}
\begin{figure}[!t]
    \centering
    \includegraphics[width=1\linewidth]{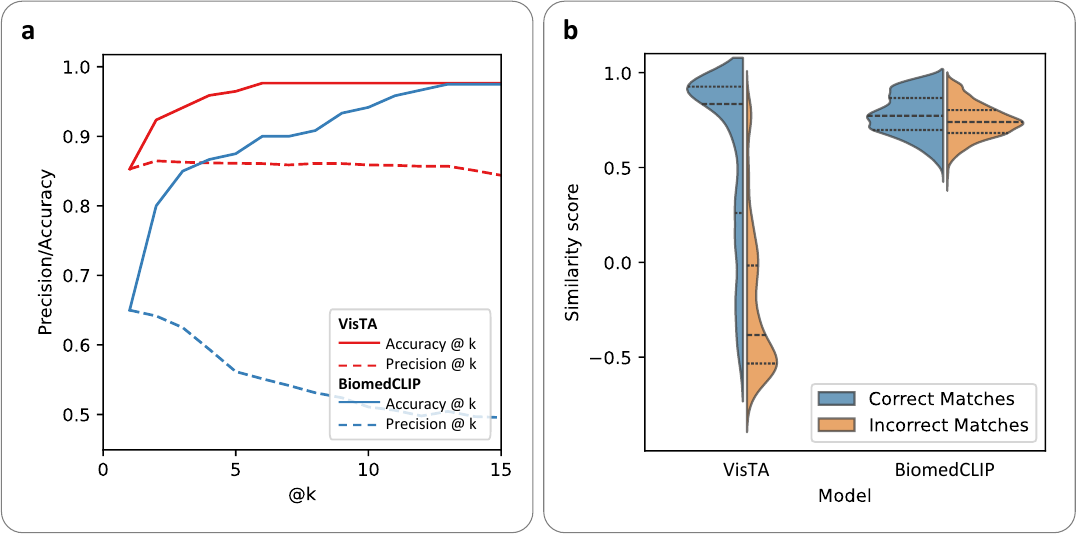}
    \caption{
        \textbf{Abnormality Retrieval Performance Analysis.}  
        The panels compare VisTA with BiomedCLIP in abnormality retrieval tasks.  
        \textbf{\textit{(a)}} Accuracy@k and Precision@k curves for top-k retrievals (k = 1 to 25). VisTA demonstrates superior retrieval accuracy and precision across k values, highlighting its effectiveness in aligning radiology images with the correct abnormality types.  
        \textbf{\textit{(b)}} Violin plots of similarity scores for correct and incorrect matches. VisTA achieves higher similarity scores for correct matches and better separation between correct and incorrect matches, demonstrating more reliable retrieval capabilities compared to BiomedCLIP.  
        These results showcase VisTA's significant improvements in abnormality retrieval, providing robust evidence for diagnostic tasks.
    }
    \label{fig:retrieval_performance}
\end{figure}

Figure~\ref{fig:retrieval_performance} evaluates the retrieval performance of VisTA compared to BiomedCLIP, focusing on \texttt{accuracy@k} and \texttt{precision@k} (Figure~\ref{fig:retrieval_performance}-a) and similarity score distributions (Figure~\ref{fig:retrieval_performance}-b).
VisTA consistently outperforms BiomedCLIP across all top-k retrievals (k = 1 to 25). It achieves higher accuracy and precision.
This indicates VisTA's superior ability to align input images with the correct abnormality types, even in top-ranked retrievals.

The violin plots in Figure~\ref{fig:retrieval_performance}-b provide additional insights into the reliability of retrievals. VisTA achieves significantly higher similarity scores for correct matches compared to BiomedCLIP, with better separation between correct and incorrect matches.
This separation reduces the likelihood of retrieving irrelevant cases and ensures that retrieved references provide meaningful evidence for diagnostic tasks.

\subsection{Reliability and Real-world Clinical Utility}

The VisTA framework and the MINDset hold promises to advance clinical decision-making by enhancing both reliability and utility in diagnostic workflows. Integrating explainable AI with robust evidence-driven reasoning, VisTA aligns with real-world clinical practices seamlessly and addresses critical gaps in current diagnostic systems.

\paragraph{Enhanced Reliability through Explainability}
VisTA's design emphasizes reliability: it provides transparent and interpretable intermediate outputs. The model retrieves reference cases with similarity scores, identifies abnormalities, and generates evidence-based explanations for each prediction. These outputs enable clinicians to verify the diagnostic reasoning process at every stage. 
For example, the refined alignment in the embedding space ensures that retrieved reference cases are contextually relevant, reducing ambiguity in diagnosis.
VisTA's ability to maintain accurate prediction performance in abnormality retrieval and diagnosis, even using limited training data (n=170), further suggests its predictions reliability.

\paragraph{Alignment with Clinical Workflows}
VisTA mirrors the multi-step clinical diagnostic process, starting from identifying abnormalities to reasoning about disease outcomes.
This modular approach allows clinicians to independently assess and validate each step, ensuring the model's outputs align with clinical expectations.
By providing explanations tied directly to retrieved references and abnormality types, VisTA enhances trust and easily integrates into existing diagnostic workflows. 
VisTAs ability to highlight subtle abnormalities and contextualize them within known cases supports clinicians in making informed decisions, especially in complex or ambiguous scenarios.

\subsection{Error Analysis}
\label{sec:error_analysis}
The error analysis sheds light on VisTA's strengths and limitations across diagnostic stages, including abnormality retrieval, reference case retrieval, and prediction tasks.
Table~\ref{tab:error_analysis} summarizes the false positive (FP) and false negative (FN) rates for each stage, alongside representative errors that illustrate common misclassifications and potential causes.

\begin{table}[!t]
\centering
\caption{\textbf{Error analysis across diagnostic stages of VisTA model using summary text modality.} Summary of false positive (FP) and false negative (FN) rates for each diagnostic stage, including abnormality retrieval, reference case retrieval, and prediction tasks. Representative errors highlight common misclassifications and potential causes.
}
\label{tab:error_analysis}
\resizebox{\textwidth}{!}{%
\begin{tabular}{llccp{.6\linewidth}}
\toprule
\multicolumn{1}{c}{\textbf{Stage}} & \multicolumn{1}{c}{\textbf{Type}} & \textbf{FP (\%)} & \textbf{FN (\%)} & \multicolumn{1}{c}{\textbf{Representative errors}} \\ \midrule
\multirow{4}{*}{\begin{tabular}[c]{@{}l@{}}\\\\\\\textbf{Abnormality}\\ \textbf{retrieval}\end{tabular}} & Normal & 50.00 & 16.67 & \begin{tabular}[c]{@{}p{\linewidth}@{}}- Misclassified as WMH due to image noise.\\ - Normal variations mistaken for subtle atrophy features.\end{tabular} \\ \cline{2-5} 
 & \begin{tabular}[c]{@{}l@{}}MTL\\ atrophy\end{tabular} & 36.00 & 27.27 & \begin{tabular}[c]{@{}p{\linewidth}@{}}- Mild atrophy misclassified as normal.\\ - Overlap with WMH features. 
 \end{tabular} \\ \cline{2-5} 
 & WMH & 0.00 & 46.15 & \begin{tabular}[c]{@{}p{\linewidth}@{}}- Missed subtle WMH regions in low-contrast images.\\ - Misclassified as MTL atrophy due to similar intensity patterns.\end{tabular} \\ \cline{2-5} 
 & \begin{tabular}[c]{@{}l@{}}Other\\ atrophy\end{tabular} & 41.67 & 46.15 & \begin{tabular}[c]{@{}p{\linewidth}@{}}- Overlap with MTL atrophy features.\\ - Subtle features mistaken for normal.\end{tabular} \\ \hline
\multicolumn{2}{l}{\textbf{Reference cases retrieval}$^*$} & \multicolumn{2}{c}{14.71} & \begin{tabular}[c]{@{}p{\linewidth}@{}}- Low-confidence cases with ambiguous similarity scores.\\ - Unique cases lacking any matching reference images.\\ - Retrieval errors for overlapping features across abnormality types.\end{tabular} \\ \hline
\multirow{2}{*}{\begin{tabular}[c]{@{}l@{}}\\\textbf{Dementia}\\ \textbf{prediction}\end{tabular}} & Dementia & 16.67 & 50.00 & \begin{tabular}[c]{@{}p{\linewidth}@{}}- WMH misclassified as non-dementia due to mild features.\\ - Subtle MTL atrophy cases missed.\end{tabular} \\ \cline{2-5} 
 & \begin{tabular}[c]{@{}l@{}}Non-\\ dementia\end{tabular} & 11.36 & 2.50 & \begin{tabular}[c]{@{}p{\linewidth}@{}}- Mild cognitive cases flagged as dementia.\\ - Normal cases with noise in images labeled incorrectly.\end{tabular} \\ \hline
\multirow{2}{*}{\begin{tabular}[c]{@{}l@{}}\\\textbf{AD}\\ \textbf{prediction}\end{tabular}} & AD & 48.28 & 37.50 & \begin{tabular}[c]{@{}p{\linewidth}@{}}- Misclassified as other dementia due to overlapping features with WMH or mild atrophy..\\ - Overemphasis on normal features in borderline cases.\end{tabular} \\ \cline{2-5} 
 & Non-AD & 42.86 & 53.85 & \begin{tabular}[c]{@{}p{\linewidth}@{}}- WMH-related non-AD misclassified as AD.\\ - Other dementia cases misclassified as AD due to feature overlap with MTL atrophy.\end{tabular} \\ \bottomrule
\multicolumn{5}{r}{\footnotesize\begin{tabular}[c]{@{}r@{}}FP (\%): Percentage of the predicted labels. FN (\%): Percentage of the true labels.\\ $^*$Incorrect matches rate of top-1 retrieval cases.\end{tabular}}
\end{tabular}%
}
\end{table}

In \textbf{abnormality retrieval}, VisTA showed strong performance in identifying normal and MTL atrophy cases.
Errors frequently arose, however, in cases with subtle features or overlapping patterns. For instance, normal cases were misclassified as WMH due to noises in the images, while mild MTL atrophy cases were misinterpreted as normal.
WMH retrieval suffered from significant false negatives, particularly in low-contrast regions where WMH features were mistakenly identified as MTL atrophy. Similarly, subtle abnormalities in ``Other Atrophy'' cases were often misclassified as normal, reflecting the challenges of detecting nuanced features.

The \textbf{reference case retrieval} stage revealed critical limitations, with $14.71\%$ of the top-1 retrieval cases failing to provide meaningful matches.
A notable example, highlighted in Case Study 3, involved a unique input image where the model's highest similarity score was only $0.52$. This low confidence score reflects a lack of reference images that are sufficiently similar in the dataset, underscoring the importance of expanding the reference dataset to include more diverse cases.
Furthermore, overlapping features across abnormality types occasionally resulted in incorrect retrievals, which emphasizes the need for better feature separation during training.

In \textbf{dementia prediction}, VisTA achieved high overall performance but encountered errors in specific cases.
WMH-related dementia cases with mild features were sometimes misclassified as non-dementia. Mild cognitive impairments in non-dementia cases were occasionally flagged as dementia due to overlapping characteristics with abnormal atrophy.
These errors highlight the challenge of distinguishing subtle abnormalities in borderline cases.

In \textbf{AD prediction}, VisTA faced challenges in distinguishing AD from other dementia types. These misclassifications often occur due to overlapping features between abnormality categories associated with AD and other forms of dementia. For instance, cases with mild MTL atrophy or WMH frequently exhibited patterns that could correspond to multiple dementia types, leading to errors in the final prediction.

This issue highlights a limitation in the final diagnosis prediction model, which requires further optimization to better handle nuanced differences between AD and other dementia types. The misclassifications suggest that the current evidence-driven reasoning process, while robust, could benefit from more specialized refinement for differentiating dementia subtypes.
The findings from the error analysis, particularly the challenges identified in Case Study 3, further emphasize areas for improvement. Enhancing the diversity and representativeness of the reference dataset, improving sensitivity to subtle features, and reducing noise-related misclassifications are key priorities for refinement.
Addressing these limitations will strengthen VisTA's performance and reliability, ensuring better alignment with clinical needs and enhancing its potential for real-world diagnostic applications.

\subsection{Limitations and Future Work}
VisTA demonstrates significant advancements in explainable AD diagnosis. Its current version has several limitations.

\textit{First}, the reference dataset, MINDset, while carefully curated and verified by medical experts, remains relatively small. This limited size may restrict the model's ability to generalize to more diverse clinical cases, especially for unique or rare abnormalities.
Expanding the dataset with additional samples and more detailed annotations for dementia subtypes could significantly enhance the model's robustness and representativeness.

\textit{Second}, the current evidence-driven reasoning process relies on fixed textual representations for abnormality types. While effective in aligning input images with reference cases, this approach may lack flexibility in handling complex or ambiguous abnormalities.
Incorporating dynamic textual representations or leveraging contextual embeddings could improve the model's ability to capture nuanced features in clinical data.

\textit{Third}, while VisTA provides strong performance in abnormality retrieval and diagnosis prediction, its ability to differentiate AD from other dementia types requires further refinement.
Future work could explore advanced training strategies, such as multi-task learning, to simultaneously optimize predictions across dementia subtypes and improve class separability.

\textit{Additionally}, the reliance on similarity scores for evidence retrieval, although clinically interpretable, may not always capture the full complexity of imaging patterns.
Combining similarity-based retrieval with more advanced reasoning frameworks, such as graph-based representations or hybrid approaches integrating other modalities (e.g., genetics or biomarkers), could further improve diagnostic accuracy and explainability.

\textit{Finally}, while VisTA aligns closely with clinical workflows, integrating the model into real-world healthcare settings will require careful validation and usability testing. Ensuring that the system is intuitive for clinicians and compatible with existing diagnostic processes will be critical for successful deployment.

\section{Conclusion}
This study introduces VisTA, a Vision-Text Alignment model, as a robust and explainable framework for AD diagnosis. VisTA bridges the gap between AI predictions and clinical interpretability by integrating radiology image analysis with evidence-driven reasoning.
Leveraging MINDset, a carefully constructed dataset of radiology images, abnormality types, and verified descriptions, VisTA ensures alignment with real-world clinical workflows and enhances diagnostic reliability.

Fine-tuned on only 170 samples, VisTA demonstrates significant improvements in both abnormality retrieval and diagnosis prediction compared to pre-trained BiomedCLIP models trained on millions of images.
This highlights the efficacy of the proposed contrastive learning approach and the quality of the MINDset. VisTA's modular architecture facilitates the retrieval of reference cases, generates detailed explanations, and provides interpretable predictions, empowering clinicians to validate intermediate results and trust final outputs.
The clear separations of clusters in the embedding spaces and retrieval performance further emphasizes the impact of contrastive learning in achieving better alignment between image and text modalities.

From an application perspective, VisTA offers transformative potential in clinical diagnostics. By providing evidence-based predictions, it aligns with multi-step diagnostic workflows, mirroring the systematic approach used by clinicians. 
This transparency fosters greater trust in AI-driven decisions and supports healthcare professionals in identifying subtle abnormalities and complex cases.
Beyond AD research, VisTA's adaptable architecture could be extended to other neurodegenerative diseases and/or imaging modalities, expanding its utility across medical domains.

In conclusion, VisTA represents a meaningful step toward explainable and reliable AI in healthcare. Integrating advanced multimodal learning with clinical reasoning demonstrates the potential of AI to enhance diagnostic accuracy, improve patient care, and align seamlessly with the needs of modern healthcare systems.
Future work will focus on expanding its applications, increasing dataset diversity, and further optimizing performance to address the ever-evolving challenges of clinical diagnostics.

\section*{Availability}

The source code for VisTA, along with the constructed MINDset, is publicly available on GitHub\footnote{\url{https://github.com/ddlinh/VisTA}}. The pre-trained VisTA model fine-tuned on the MINDset are available on Hugging Face\footnote{\url{https://huggingface.co/ddlinh/vista}}.
The publicly available Alzheimer's Disease prediction dataset utilized in this study are accessible via the Hugging Face Datasets platform\footnote{\url{https://huggingface.co/datasets/Falah/Alzheimer_MRI}}.

\section*{Declaration of Competing Interest}
The authors declare that they have no known competing financial interests or personal relationships that could have appeared to influence the work reported in this paper.

\section*{Declaration of generative AI and AI-assisted technologies in the writing process}
During the preparation of this work, D.C. Can used ChatGPT to improve the language of the work. After using this tool/service, D.C. Can reviewed and edited the content as needed, and took full responsibility for the content of the published article.

\section*{Funding}
This research did not receive any specific grant from funding agencies in the public, commercial, or not-for-profit sectors.

\section*{Acknowledgments}
Some of the graphics used in preparing the figures are designed by Freepik (\url{www.freepik.com}).





\bibliographystyle{elsarticle-num} 
\bibliography{refs}






\end{document}